# Behavior Cloning of MPC for 3-DOF Robotic Manipulators

Theo Guegan
*University of Waterloo*
theo.guegan@etu.utc.fr

Wen Jie Dexter Teo
*University of Waterloo*
WTEO030@e.ntu.edu.sg

***Abstract*—While Model Predictive Control (MPC) provides strong stability and robustness, it imposes a significant computational burden on real-time systems. This paper investigates the application of Behavior Cloning to approximate MPC policies for the real-time control of a 3-degree-of-freedom robotic manipulator. We present a baseline controller combining Inverse Kinematics with MPC and evaluate neural network architectures, ranging from classical regression algorithms to deep learning models including Deep MLPs and RNNs, to derive computationally efficient surrogate policies. We analyze generalization capabilities, stability considerations, and the trade-offs inherent in different architectural choices. Our empirical study employs both online and offline evaluations to assess performance regarding accuracy, computational efficiency, and fidelity to the original MPC policy. Our results demonstrate that Behavior Cloning can effectively reduce the computational burden of MPC policies for 3-DOF robotic manipulators, achieving a 3x reduction in inference latency with a 84.98% success rate under relaxed tolerances. Notably, we find that static architectures outperform temporal variants, confirming the sufficiency of instantaneous state observations for this task. However, we observe a precision gap under strict tolerances, which suggest that while Behavior Cloning captures the global optimal trajectory, further research is needed to minimize terminal steady-state error.**

## I. Introduction

Model Predictive Control (MPC) has been widely used for robotic manipulation [1], offering an optimal control strategy with strong stability and robustness. However, the computational cost of MPC for solving the optimization problems limits its applicability for both real-time systems and resource-constrained devices. Neural networks, with their diverse architectures, offer a promising and computationally efficient alternative for approximating MPC policies [2]. We consider a 3-degree-of-freedom (3-DOF) robotic manipulator operating in a MuJoCo simulation environment. The simulation environment provides a realistic and controllable environment for testing and evaluating the proposed methodology. MuJoCo also handles gravity compensation and joint friction, allowing us to simplify the control problem and focus on the learning aspect. The control objective centers on driving the end-effector to reach a 3D cartesian target position within the robot's reachable workspace. Inspired by the recent usage of imitation learning for complex controls [3], we present a complete data generation pipeline for collecting high-quality demonstrations of the desired behavior and an empirical evaluation of both feedforward and recurrent neural networks for policy learning. Our experiment focuses on minimizing the control error and testing the ability of the learned policy to generalize in the simulation environment.

[1] This work was conducted during an exchange program at University of Waterloo. T. Guegan is currently with the Universite de Technologie de Compiègne, France, and ML Intern at Opalin, Redwood City, CA. W. J. D. Teo is currently with Nanyang Technological University, Singapore, and Research Intern at Polytechnique Montréal, QC, Canada.
Code available : https://github.com/theguega/neural-mpc

## II. Problem Formulation

### A. System Description

We consider a 3-DOF robotic manipulator defined by generalized coordinates $q = [q_1, q_2, q_3]^T \in \mathbb{R}^3$, representing joint angles, and their time derivatives $\dot{q} \in \mathbb{R}^3$. The full observable state at discrete time step $k$ is

$$x_k = \left[q_k^T, \dot{q}_k^T\right]^T \in \mathbb{R}^6 \tag{1}$$

The manipulator operates in a MuJoCo simulation environment (Figure 2) governed by rigid-body dynamics with gravity compensation. The control objective is to drive the end-effector to track randomly sampled, reachable 3D Cartesian target positions $p_{\text{des}} \in \mathbb{R}^3$ within the robot's workspace $\mathcal{W} \subset \mathbb{R}^3$.

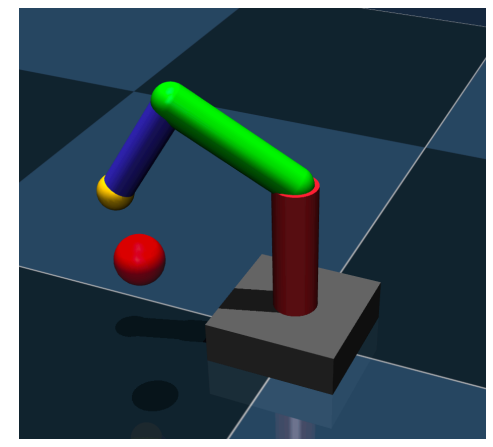
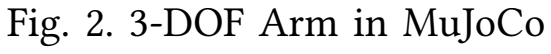

Fig. 2. 3-DOF Arm in MuJoCo

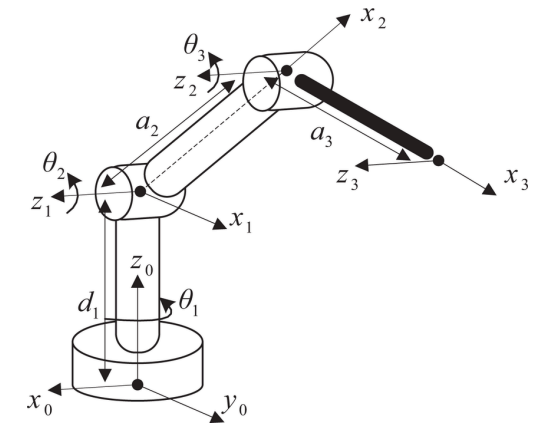

Fig. 3. 3-DOF Arm Schema [4]

## III. Baseline Controller: MPC with Inverse Kinematics

Our baseline controller uses a hierarchical architecture combining an Inverse Kinematics (IK) module and a MPC module. The IK module computes the joint angles required to achieve the desired end-effector position, while the MPC module optimizes the joint velocities to minimize the control error.

### A. Inverse Kinematics Formulation

The IK module translates desired end-effector positions into feasible joint-space configurations. Let $p(q) : \mathbb{R}^3 \to \mathbb{R}^3$ denote the forward kinematics mapping. The Cartesian error is defined as:

$$e = p_{\text{des}} - p(q) \tag{2}$$

We solve the IK problem using the Jacobian transpose method with Damped Least Squares (DLS) for numerical stability near singularities. The iterative update rule is:

$$\Delta q = J^T \left(JJ^T + \lambda^2 I\right)^{-1} e \tag{3}$$

where $J(q) = \mathrm{d}(\partial p, \partial q) \in \mathbb{R}^{3\times3}$ denotes the geometric Jacobian and $\lambda$ represents the damping factor.

To prevent overshooting or divergence, the joint update is clamped to a maximum norm relative to the step size $\alpha \in [0, 1]$:

$$\Delta q = \begin{cases} \Delta q \text{ if } \|\Delta q\| \leq \alpha \\ \Delta q * \frac{\alpha}{\|\Delta q\|} \text{ otherwise} \end{cases} \tag{4}$$

Finally, joint angles are wrapped to avoid numerical drift:

$$q_i \leftarrow \text{atan2}(\sin(q_i), \cos(q_i)) \tag{5}$$

### B. Model Predictive Control Formulation

The MPC module is given the desired joint angles $q_{\text{des}} \in \mathbb{R}^3$ from the IK module and computes optimal control torques $\tau_{\text{MPC}}$ with a specified prediction horizon. We simplify the system dynamics for the MPC formulation by assuming unit inertia and neglecting Coriolis effects, justified by MuJoCo's compensation of complex dynamics. This yields a double-integrator model where the control input $\tau_{\text{MPC}}$ effectively commands joint acceleration:

$$\ddot{q} = \tau_{\text{MPC}} \tag{6}$$

With $x = [q, \dot{q}]^T \in \mathbb{R}^6$ and discrete-time dynamics:

$$x_{k+1} = x_k + \Delta t * \begin{bmatrix} \dot{q}_k \\ \tau_{\text{MPC,k}} \end{bmatrix} = f(x_k, \tau_k) \tag{7}$$

The MPC solves a finite-horizon optimal control problem with quadratic cost function:

$$\min_{\tau_{0:N-1}} \sum_{k=0}^{N-1} \left( \|x_k - x_{\text{ref}}\|_{\text{Q}}^2 + \|\tau_k\|_{\text{R}}^2 \right) + \|x_N - x_{\text{ref}}\|_{\text{Q}_\text{N}^2} \tag{8}$$

subject to:

$$x_{k+1} = f(x_k, \tau_k), \quad x_0 = x(t), \quad \tau_{\min} \leq \tau_k \leq \tau_{\max} \tag{9}$$

Where $x_{\text{ref}} = \left[q_{\text{des}}, 0^T\right]^T$ is the target state, and $\boldsymbol{Q}, \boldsymbol{R}, \boldsymbol{Q_N}$ are positive definite matrices. The optimization problem is solved using CasADI [5] with OSQP optimization solver.

## IV. Data Generation Pipeline

To enable behavior cloning from the expert IK–MPC controller, we generate a dataset of joint angles, joint velocities, target positions, and predicted torques from a closed-loop MuJoCo simulation. The process for each episode is:

### A. Collection process

1) Target sampling: A reachable end-effector target $p_{\text{des}} \in \mathbb{R}^3$ is sampled within the workspace $\mathcal{W}$ using cylindrical coordinates to sample a radius $r$ and height $z$ uniformly within the workspace bounds.
2) The IK solver computes the corresponding joint-space reference $q_{\text{des}}$.
3) Given the current state $x_k$ and a specified prediction horizon $N$, the MPC controller generates torque commands $\tau_{\text{MPC}}$ to achieve the desired joint angles and velocities.
4) For each time step $k$, we record the current state $[q_1, q_2, q_3, \dot{q}_1, \dot{q}_2, \dot{q}_3] \in \mathbb{R}^6$, the target $p_{\text{des}} \in \mathbb{R}^3$, and the predicted torque $\tau_{\text{MPC}} \in \mathbb{R}^3$.
5) We step the simulation until the end of the episode or until the target is reached using $\tau = \tau_{\text{MPC}} + \tau_{\text{env}}$ (where $\tau_{\text{env}}$ represents the forces computed by MuJoCo to compensate for Coriolis, centrifugal, and gravitational effects, ensuring the simplified dynamics model in the MPC controller remains valid; MuJoCo bias force: mjData.qfrc_bias).

During this process, if either the MPC controller or the IK solver fails to converge, we discard the data for that time step to retain only high-quality samples.

### B. Dataset Structure

After generation, the data is stored in HDF5 format, grouped by episodes containing synchronized sequences of states ($6D$), targets ($3D$), and actions ($3D$).

This hierarchical format preserves the temporal integrity of each trial, allowing us to process the data differently depending on the model architecture. The raw data is loaded via a custom MPCDataset class, which constructs the input feature vector $x$ by concatenating the state ($\mathbb{R}^6$) and the target ($\mathbb{R}^3$), resulting in a 9-dimensional input vector.

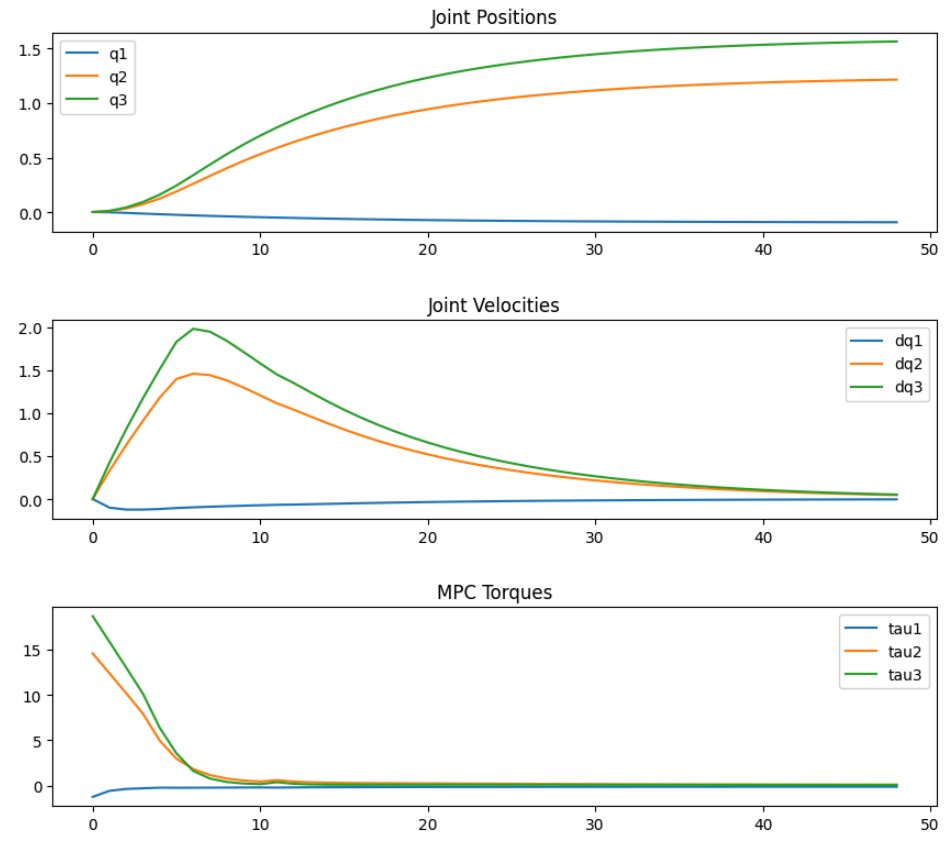

Fig. 4. Visualization of a random episode

Depending on the learning algorithm, the data is processed differently according to the model architecture.

a) *Flat Formatting:*

For non-sequential algorithms (e.g., MLPs, Random Forests), temporal dependencies are discarded to maximize sample efficiency. We treat every timestep $t$ from every episode as an independent sample (i.i.d).

$$X_{\text{flat}} \in \mathbb{R}^{N\times9}, \quad Y_{\text{flat}} \in \mathbb{R}^{N\times3} \tag{10}$$

Where $N = \sum_{i=0}^{E} T_i$ is the total number of timesteps across all episodes.

b) *Sequential Formatting:*

For time-series algorithms such as GRU, preserving the temporal dependencies is crucial. We treat every episode as a sequence of timesteps, where each timestep is a sample.

$$X_{\text{seq}} \in \mathbb{R}^{E \times T \times 9}, \quad Y_{\text{seq}} \in \mathbb{R}^{E \times T \times 3} \tag{11}$$

Where $E$ is the number of episodes and $T$ is the number of timesteps per episode.

c) *Sliding Window Formatting:*

To approximate a recurrent structure with our MLP architecture, we augment each time step with its previous $W$ timesteps. This injects short-term memory into an otherwise i.i.d. formulation.

For each timestep $t$, we create a new input vector $X_{\text{seq}_{\{t\}}}$ by concatenating the current state $X_{\{t\}}$ with the previous $W$ states $X_{\{t-1\}}, X_{\{t-2\}}, ..., X_{\{t-W\}}$, and the target $\in R^3$ only once to avoid redundancy.

$$X_{\text{seq}} \in \mathbb{R}^{N \times (W \cdot 6+3)}, \quad Y_{\text{seq}} \in \mathbb{R}^{N} \tag{12}$$

## C. Data Preprocessing

Our goal is to develop a robust and reliable controller that can handle uncertainties and disturbances in the system. To this end, we introduce small Gaussian noise to both the input state $[q_1, q_2, q_3, \dot{q}_1, \dot{q}_2, \dot{q}_3]$ and the output action $\tau_{\text{MPC}}$. This noise simulates real-world conditions such as sensor noise, actuator noise, and environmental disturbances. Because the data generation pipeline can produce an arbitrary number of samples, we collect a large dataset for training and increase the number of samples until the validation loss plateaus or computational limits are reached.

We ran a simple experiment with both SVR and MLP models from Scikit-learn and compared their performance while increasing the number of samples progressively.

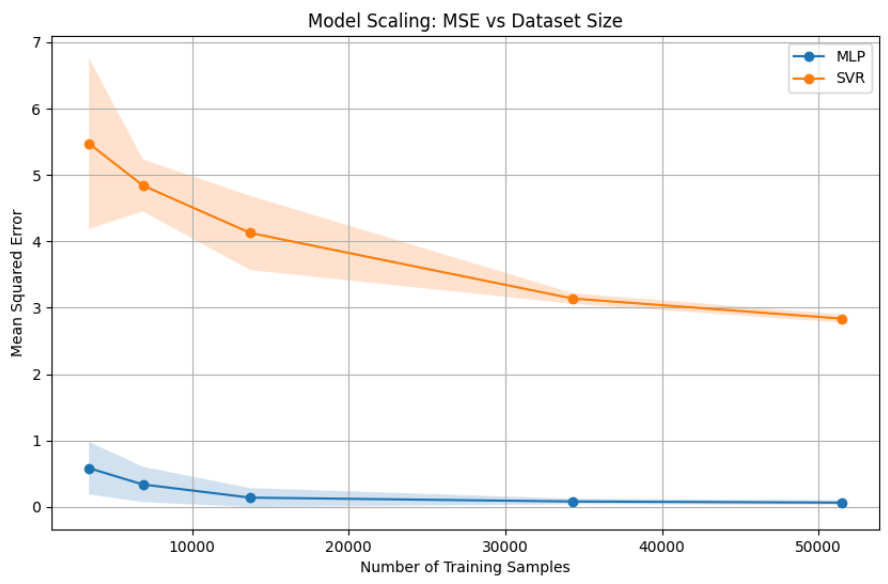

Fig. 5. MSE vs number of samples (5 trials per model)

Our experiment showed a clear plateau after 30000 samples Figure 5, therefore we decided to use only 35000 samples for the training of our regression algorithms. However, we also observed that RNN models such as GRU were taking full benefits of the full dataset by improving significantly compared to MLP models even after 30000 samples.

Nevertheless, since our focus is on finding a suitable architecture for a neural surrogate, we continue to utilize the full dataset for training all of our models for a fair comparison.

# V. Neural Network Architecture

We formulated the learning problem as a regression task, where the goal is to predict the torque $\tau_{\text{MPC}}$ given the current state $x_k$ and the target $p_{\text{des}}$. We aimed to minimize the error between the neural network policy $\pi_\theta$ and the expert MPC actions:

$$\min_{\theta} \quad L\left(\pi_{\theta(X)}, \tau_{\text{MPC}}\right) \tag{13}$$

where $L$ is a loss function that measures the difference between the predicted torque and the expert MPC torque. We investigated a range of models, from traditional machine learning to deep learning architecture, to understand their effectiveness in approximating the MPC policy. For this task, we compared the performance with 2 different loss functions:

- Mean Squared Error (MSE)
- Mean Absolute Error (MAE)

## A. Regression Baselines

To establish performance benchmarks, we evaluated several standard regression algorithms from the Scikit-learn library. These models operate on the "flat" dataset format, treating each timestep as an independent sample. We included tree-based regressors (Random Forest and Gradient Boosting) and a shallow MLP regressor as a lightweight baseline to compare against our deeper custom architectures.

## B. Custom Multi-Layer Perceptron (MLP)

We implemented a custom feedforward network to explore the impact of model capacity on cloning accuracy. This model processes the flat input vector through a series of fully connected linear layers with ReLU activations. We conducted an architectural search by varying:

- Depth: Number of hidden layers
- Width: Number of neurons per layer

This memory-less architecture captures and learns directly the mapping from the current state and target to the required control action.

## C. Time Series Models

To investigate whether historical context improves control fidelity, we evaluated architectures designed to capture temporal dependencies.

a) *Sliding Window MLP:*

We implemented Sliding Window variants with the same depth and width parameters as our custom MLP. These variants receive a concatenated history of the past W state observations as input, where we set $W = 5$. This explicitly injects short-term memory into the network to test if providing temporal history can improve predictions.

b) *Recurrent Neural Networks (RNNs):*

We evaluated Gated Recurrent Units (GRU) networks. These models maintain a hidden state $h_t$ that summarizes the history of the episode up to time $t-1$. The final hidden state is then passed through a linear output layer to produce the predicted torque $\pi_{\theta(X)}$. Here again, we varied the number of hidden units per layer and the number of layers.

$$h_t = \text{RNN}(x_t, h_{t-1}), \quad \pi_{\theta(X)} = \text{Linear}(h_t) \tag{14}$$

## VI. Evaluation Methodology

We evaluated the learned policies using a combination of offline and online metrics:

### A. Offline Metrics

We used MAE and Root Mean Squared Error (RMSE) to measure the average deviation of the predicted torques from the expert torques. Additionally, we defined Direction Accuracy (DA) to evaluate whether the model correctly identifies the sign (direction) of joint acceleration.

$$\mathrm{DA} = \frac{1}{3N}\sum_{i=0}^{N}\sum_{j=1}^{3}\mathbb{I}\Big(\mathrm{sign}\big(\tau_{\mathrm{MPC,i,j}}\big) = \mathrm{sign}\Big(\pi_\theta(X_i)_j\Big)\Big) \quad (15)$$

Finally, we measured the proportion of variance in the expert's action that is explained by our model, using explained variance. It is a normalized, scale-invariant metric for comparing performance defined as follows:

$$\text{Explained Variance} = 1 - \frac{\mathrm{Var}\big(\tau_{\mathrm{MPC}} - \pi_{\theta(X)}\big)}{\mathrm{Var}(\tau_{\mathrm{MPC}})} \quad (16)$$

### B. Online Metrics

To assess the closed-loop performance, we deployed the trained policies in our simulation environment and evaluated them on the following criteria:

- Success Rate: The percentage of episodes where the end-effector's final position converges within a specified tolerance $\varepsilon$ of the target
$$\|p_{\mathrm{final}} - p_{\mathrm{des}}\|_2 < \varepsilon \quad (17)$$
- Average Position Error: The mean Euclidean distance between the end-effector and the target across the entire trajectory. This verifies that the model actively minimizes error rather than just drifting near the goal.
- Computational Time Efficiency: The average inference latency per control step. We compare this against the baseline MPC solution time to confirm that the neural networks achieve higher control frequencies.
- Computational Cost: Average CPU utilization during operation, ensuring the surrogate model is sufficiently lightweight for potential deployment on embedded systems.

## VII. Results

### A. Regression Baseline

For our regression baseline with scikit-learn, we evaluated several standard regression algorithms using the collected offline dataset. As discussed before, we used only a subset of the whole dataset for a total of 35000 samples, split into 80% for training and 20% for testing. The input feature space $X \in \mathbb{R}^9$ consists of the robot's current state and target coordinates, while the output target $Y \in \mathbb{R}^3$ corresponds to the applied joint actions (torques). The data was partitioned into a training set (80%) and a test set (20%) via random shuffling. To ensure the statistical significance of the reported metrics, each model was trained and evaluated over 5 independent runs. Table I summarizes the performance across MSE, MAE, Explained Variance, and Directional Accuracy. We also tried scaling our features using Scikit-Learn's StandardScaler, which did not significantly improve performance.

TABLE I
Comparison of regression algorithms on the validation set (averaged over 5 runs).

| Model | MSE (Mean ± CI) | MAE (Mean ± CI) | Expl. Var | Dir. Acc. | MSE/ Torque |
|---|---|---|---|---|---|
| **Ridge** | 8.729 ± 0.098 | 1.418 ± 0.009 | 0.221 | 0.681 | [4.24, 5.54, 16.41] |
| **Random Forest** | 0.097 ± 0.003 | 0.111 ± 0.002 | 0.991 | 0.999 | [0.05, 0.08, 0.16] |
| **MLP Regressor** | 0.053 ± 0.013 | 0.094 ± 0.009 | 0.994 | 0.938 | [0.05, 0.03, 0.07] |
| **Gradient Boosting** | 1.017 ± 0.059 | 0.243 ± 0.005 | 0.843 | 0.996 | [2.30, 0.10, 0.65] |
| **KNN Regressor** | 0.237 ± 0.038 | 0.091 ± 0.002 | 0.976 | 0.999 | [0.20, 0.11, 0.40] |

The results highlight the inherent non-linearity of the inverse dynamics mapping. Linear Regression failed to capture the underlying relationship, exhibiting high variance across all torque dimensions. In contrast, non-linear methods performed significantly better. The MLP Regressor achieved the lowest overall MSE (0.053), indicating its superior capability in minimizing large control deviations, which is critical for preventing hardware damage. While KNN Regressor achieved the highest Directional Accuracy (99.87%) and lowest MAE, its higher MSE suggests it suffers from occasional large prediction errors (outliers). Finally, different SVM models with linear and radial basis function (RBF) kernels were evaluated. However, as mentioned in the documentation, RBF kernels cannot scale with that many samples, whereas linear kernels yielded poor performance.

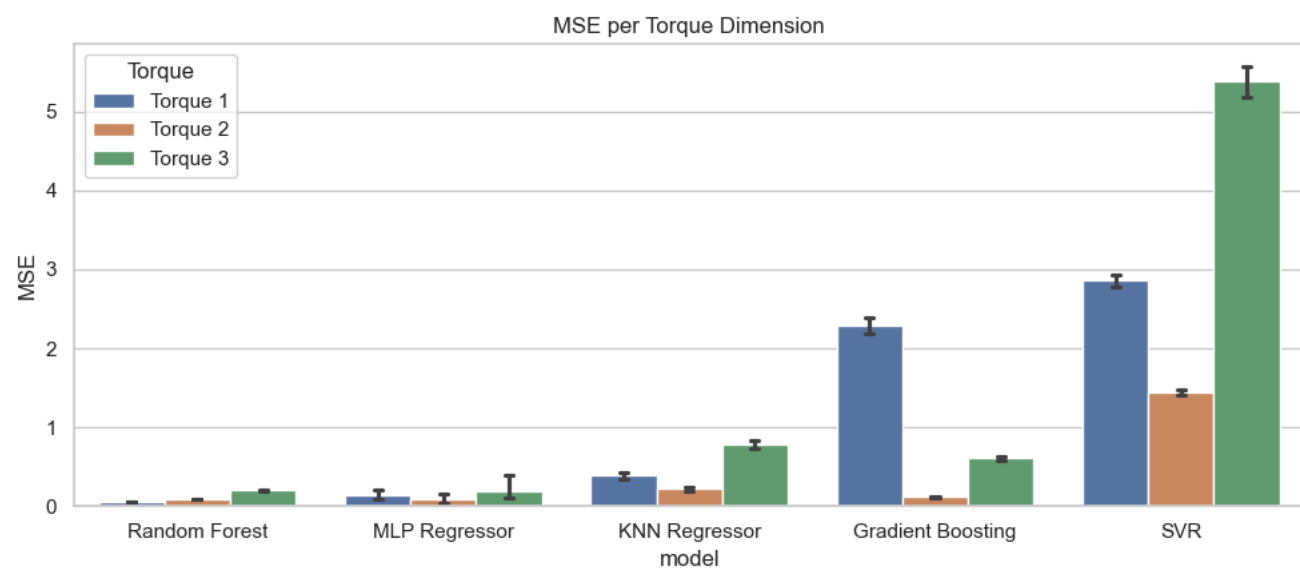

Fig. 6. Mean Squared Error per Torque

### B. Loss comparison

In this section, we compare performance of training models with two different loss functions. For this part we used a PyTorch implementation of the MLP regressor and of a GRU

RNNs. Table II summarizes the results of our experiments where the name of the model configuration is the name of the model and the loss function used for training.

TABLE II
Loss Comparison

| Model Config | MSE (Mean ± Std) | MAE (Mean ± Std) |
|---|---|---|
| **MLP_mse** | $0.1460 \pm 0.0522$ | $0.1474 \pm 0.0126$ |
| **MLP_mae** | $0.2017 \pm 0.0455$ | $0.0892 \pm 0.0078$ |
| **GRU_mse** | $1.1442 \pm 0.1441$ | $0.2581 \pm 0.0138$ |
| **GRU_mae** | $2.1821 \pm 0.1001$ | $0.3570 \pm 0.0101$ |

Following this experiment, we adopted the MSE loss function for our experiments because models trained with it consistently achieved lower test error and lower variance across runs, as shown in Table II. In particular, the MLP_mse and GRU_mse configurations outperformed their MAE-trained counterparts in terms of MSE, which was our primary performance metric for policy imitation.

### C. Hyperparameters Tuning of MLP and GRU

In this section, we present the offline evaluation metrics collected during the training of MLP and GRU models to determine the optimal architecture for our task.

a) *Experimental Setup:* To ensure the reliability of our results, each model configuration was trained and evaluated 5 times. The dataset was split into training (80%), validation (10%), and testing (10%) sets. The results presented in Figure 7 reflect the MSE on the test set.

We also monitored validation loss to detect potential overfitting. No significant overfitting was observed in any of the experiments across the tested architectures and data augmentation strategies. As established in previous sections, the MLP models were trained on a subset of 35,000 timesteps, which was deemed sufficient for convergence, while the GRU models utilized the full dataset to capture temporal dependencies effectively.

b) *Tested Architectures:* We evaluated four variations of the Multi-Layer Perceptron (MLP) and four variations of the Gated Recurrent Unit (GRU) network. The specific hyperparameters for each configuration are detailed in Table III.

TABLE III
Summary of model architectures and hyperparameters used during tuning.

| Model Name | Type | Parameters |
|---|---|---|
| MLP_Small | MLP | Hidden Layers: [64, 32] |
| MLP_Medium | MLP | Hidden Layers: [128, 64] |
| MLP_Deep | MLP | Hidden Layers: [256, 128, 64, 32] |
| MLP_Deep_Scaled | MLP | Hidden Layers: [512, 256, 128, 64] |
| GRU_Shallow | GRU | Hidden Dim: 64, Layers: 1 |
| GRU_Medium | GRU | Hidden Dim: 128, Layers: 2 |
| GRU_Deep | GRU | Hidden Dim: 128, Layers: 4 |
| GRU_Wide | GRU | Hidden Dim: 256, Layers: 2 |

To approximate temporal awareness, we also implemented Sliding Window variants (W=5) across the four MLP architectures, using the same hidden layer parameters.

c) *Results Analysis:*

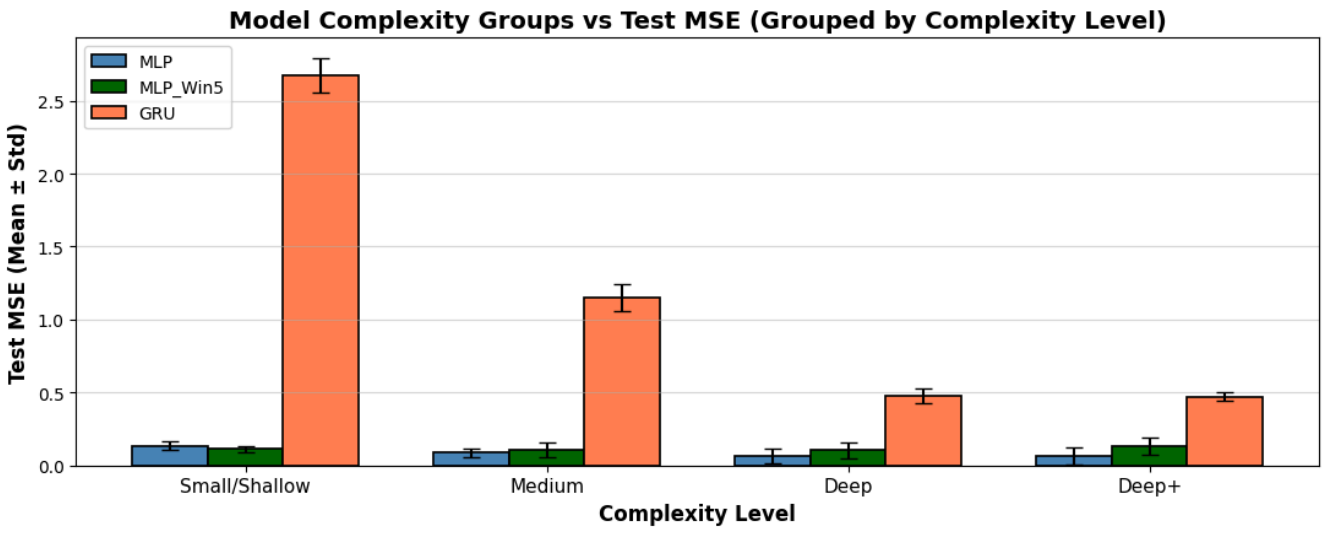

Fig. 7. Architecture Comparison (Metric: MSE). Error bars represent the standard deviation across 5 runs. Note: GRU_Wide is plot under Deep+.

The performance comparison in Figure 7 highlights distinct trends between the MLP and GRU architectures:

1) **MLP Superiority:** In our experimental setting, the MLP architectures consistently outperformed the GRU variants. The MLP_Deep configuration achieved the lowest Mean MSE overall (approximately 0.05).
2) **Depth vs. Width:** For the MLP, increasing network depth provided significant performance gains, with MLP_Deep notably outperforming MLP_Medium and MLP_Small. Increasing width did not improve performance as MLP_Deep_Scaled was unable to achieve a significant reduction in MSE over MLP_Deep. This suggests that for this task, increasing the depth of the network is more beneficial for performance.
3) **Impact of Sliding Window:** Contrary to the expectation that history would aid prediction, the inclusion of a sliding window did not result in a reduction in MSE. In fact, the sliding window resulted in a slight increase in MSE for medium and deep models. This counter-intuitive result suggests that for this specific task, the current state contains sufficient information to determine the optimal control action (Markov property). The sliding window likely complicated the optimization landscape, introducing redundant noise to the model.
4) **GRU Performance:** The GRU models struggled to match the precision of the MLPs. GRU_Shallow performed the worst among all tested configurations (MSE $\approx 2.7$). Increasing complexity helped; GRU_Deep and GRU_Wide achieved significant improvements upon the shallower variants, though still lagging behind the MLP models.

Based on these results, MLP_Deep demonstrates the strongest predictive capability and stability on the test set.

### D. Online evaluation (MuJoCo)

To validate these offline findings, we conducted a comprehensive online evaluation. We deployed every trained controller into the closed-loop MuJoCo simulation. These were benchmarked against the original MPC policy (Expert). We executed 1000 random test episodes, with a limit of 150 steps per episode, for each model. Performance was measured using **Success Rate**, defined as the percentage of episodes where the end-effector converges within a Euclidean distance $\varepsilon$ of the target. To analyze precision, three tiers were established: Strict ($\varepsilon = 0.02m$), Moderate ($\varepsilon = 0.03m$), and Relaxed ($\varepsilon = 0.05m$). We also tracked **Inference Latency** to quantify the speedup relative to the MPC solver.

a) *Success Rate Results:* While Scikit-learn baselines failed to control the robot ($< 35\%$ success rate), the PyTorch-based architectures produced results comparable to the MPC.

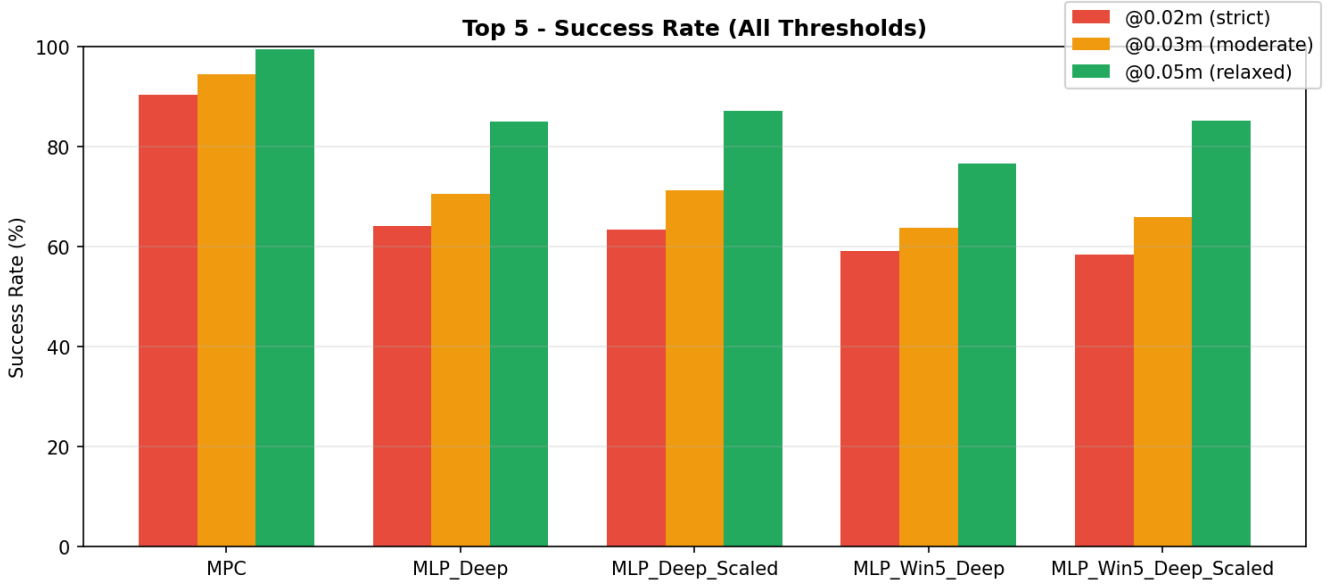

Fig. 8. Comparison of closed-loop success rates across different error tolerances ($\varepsilon$). The bar graphs represent the mean values across 5 runs.

Figure 8 illustrates the performance of the top-performing models. The results highlight a precision gap in behavior cloning. MLP_Deep was able to achieve 84.98% success rate under relaxed tolerances, but only 64.16% under strict tolerance. However, the mean final tracking error was only 2.9 cm, confirming that the majority of failures were near-misses. This evaluation also confirms our results from earlier, whereby static MLP outperformed temporal architectures, validating that temporal history is unnecessary for this task.

b) *Computational Efficiency:* We compared the distribution of solve times between the MPC and MLP_Deep.

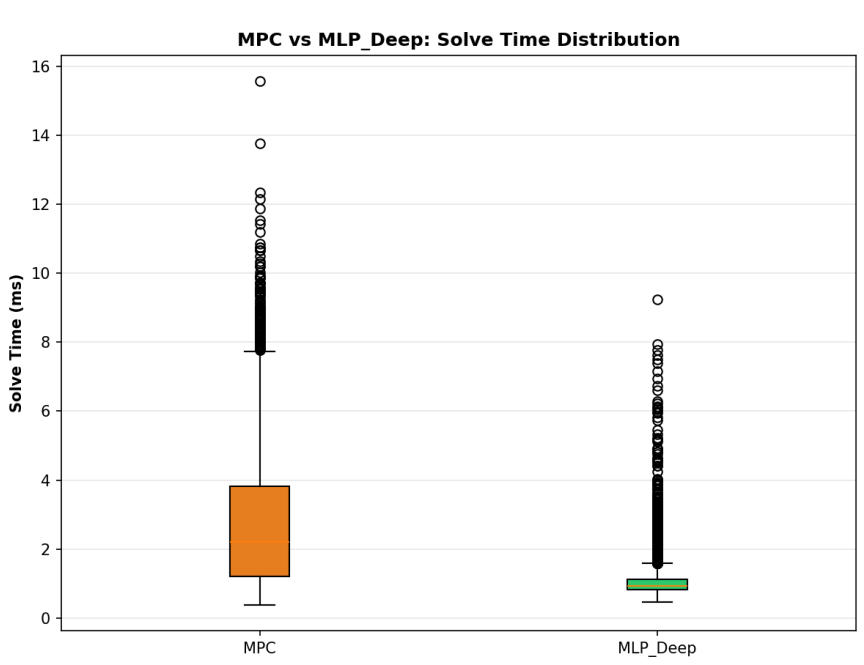

Fig. 9. Distribution of Inference Latency per Control Step.

The MLP_Deep policy achieved a mean solve time of 1.102 $\pm$0.614ms. This is nearly one-third of the mean solve time of the MPC Expert. Notably, the neural network offers a much more deterministic inference time as seen in Figure 9, making it more suitable for hard real-time constraints than the MPC, whose solve time fluctuates significantly based on the optimization landscape. Additionally, MLP_Deep utilized significantly fewer system resources, with a 25% decrease in CPU utilization as compared to the MPC.

### E. *Discussion and Conclusion*

Our results show a strong alignment between offline regression metrics and closed-loop control performance. The low offline MSE $\approx$ 0.5 successfully translated to a robust online policy, achieving a 84.98% success rate under relaxed tolerances. The mean final tracking error of 2.9cm is not a failure of generalization, but rather a direct reflection of the resolution limit inherent in the offline data. The neural network successfully learned the expert's global trajectory, but lacks the gradient-based feedback required to eliminate the final centimeter of error.

Crucially, the failure of Sliding Window and GRU models to outperform static MLPs confirms that the dynamics for our specific setup are strictly Markovian and are fully captured by the current state $(q, \dot{q})$. Regarding model capacity, while network depth proved critical for capturing the non-linear inverse dynamics, increasing width did not achieve significant performance gains. This suggests that for this specific task, adding more layers is far more effective than adding more neurons.

In conclusion, a standard Feedforward Neural Network is able to approximate an MPC policy with approximately 1ms inference latency and 2.9cm average error. MLP_Deep emerged as the optimal architecture, achieving significant improvements in speed and resource utilization, confirming that the learned policy is sufficiently lightweight for deployment on resource-constrained embedded systems.

## VIII. Future Work

This work establishes a foundation for behavior cloning of MPC on 3-DOF manipulators, which can be extended in several directions. Firstly, to bridge the precision gap under strict tolerances, future work should incorporate Data Aggregation (DAgger), allowing the learner to query the expert and correct the terminal steady-state error.

Secondly, future work should investigate this approach on robotic manipulators with higher degrees of freedom (e.g., 6-DOF). We hypothesize that increasing network depth and dataset size will become critical when state and action space dimensionality increases significantly.

Additionally, to advance towards real-world deployment, the methodology should be extended to handle more complex control scenarios. It could be interesting to investigate the cloning of a non-linear MPC which is capable of handling more complex dynamics.

From a methodological perspective, exploring advanced neural network architectures represents a promising direction. Transformer models, with their self-attention mechanisms, could be investigated for their ability to capture complex, long-range dependencies. Furthermore, the Legendre Memory Unit (LMU) [6], developed at the University of Waterloo, offers a complementary, principled approach to continuous time memory, which may prove to be well-suited for the robotic system's underlying dynamics. Inverse reinforcement learning [3] may prove to be an efficient alternative to learn the underlying MPC cost function.

## Acknowledgments

This project is a final project for the course “Foundations of Artificial Intelligence” - SYDE522. We would like to thank our instructor Terry Stewart for his guidance and support.

## References

[1] Z. Zhou, Y. Zhang, and Y. Li, “Model Predictive Control Design of a 3-DOF Robot Arm Based on Recognition of Spatial Coordinates.” [Online]. Available: https://arxiv.org/abs/2209.01706

[2] C. Gonzalez, H. Asadi, L. Kooijman, and C. P. Lim, “Neural Networks for Fast Optimisation in Model Predictive Control: A Review.” [Online]. Available: https://arxiv.org/abs/2309.02668

[3] V. G. de A. Porto, D. C. Melo, M. R. Maximo, and R. J. Afonso, “Imitation learning of a model predictive controller for real-time humanoid robot walking,” *Engineering Applications of Artificial Intelligence*, vol. 143, p. 109919, Mar. 2025, doi: 10.1016/j.engappai.2024.109919.

[4] N. Ngoc Son, H. P. H. Anh, and N. Thanh Nam, “Robot manipulator identification based on adaptive multiple-input and multiple-output neural model optimized by advanced differential evolution algorithm,” *International Journal of Advanced Robotic Systems*, vol. 14, no. 1, Dec. 2016, doi: 10.1177/1729881416677695.

[5] J. A. E. Andersson, J. Gillis, G. Horn, J. B. Rawlings, and M. Diehl, “CasADi – A software framework for nonlinear optimization and optimal control,” *Mathematical Programming Computation*, 2018.

[6] A. Voelker, I. Kajić, and C. Eliasmith, “Legendre Memory Units: Continuous-Time Representation in Recurrent Neural Networks,” in *Advances in Neural Information Processing Systems*, H. Wallach, H. Larochelle, A. Beygelzimer, F. d'Alché-Buc, E. Fox, and R. Garnett, Eds., Curran Associates, Inc., 2019, p. . [Online]. Available: https://proceedings.neurips.cc/paper_files/paper/2019/file/952285b9b7e7a1be5aa7849f32ffff05-Paper.pdf